# Integrating Large Language Models with Human Expertise for Disease Detection in Electronic Health Records

Jie Pan[1,2,4,#], PhD; Seungwon Lee[1,2,3], MPH PhD; Cheligeer Cheligeer[1,3], PhD; Elliot A. Martin[1,3], PhD; Kiarash Riazi[1,2,4], MD; Hude Quan[1,2,4], MD PhD; Na Li[1,2,4], PhD

[1] Centre for Health Informatics, Cumming School of Medicine, University of Calgary, Calgary, AB, Canada

[2] Department of Community Health Sciences, Cumming School of Medicine, University of Calgary, Calgary, AB, Canada

[3] Provincial Research Data Services, Alberta Health Services, Calgary, AB, Canada

[4] Libin Cardiovascular Institute, University of Calgary, Calgary, AB, Canada

\# Corresponding author

Jie Pan, PhD

Department of Community Health Sciences

Cumming School of Medicine

University of Calgary

CWPH 5E09 3280 Hospital Dr NW

Calgary, AB T2N 4Z6, Canada

Email: jie.pan@ucalgary.ca

# ABSTRACT

**Objective**

Electronic health records (EHR) are widely available to complement administrative data-based disease surveillance and healthcare performance evaluation. Defining conditions from EHR is labour-intensive and requires extensive manual labelling of disease outcomes. This study developed an efficient strategy based on advanced large language models to identify multiple conditions from EHR clinical notes.

**Methods**

We linked a cardiac registry cohort in 2015 with an EHR system in Alberta, Canada. We developed a pipeline that leveraged a generative large language model (LLM) to analyze, understand, and interpret EHR notes by prompts based on specific diagnosis, treatment management, and clinical guidelines. The pipeline was applied to detect acute myocardial infarction (AMI), diabetes, and hypertension. The performance was compared against clinician-validated diagnoses as the reference standard and widely adopted International Classification of Diseases (ICD) codes-based methods.

**Results**

The study cohort accounted for 3,088 patients and 551,095 clinical notes. The prevalence was 55.4%, 27.7%, 65.9% and for AMI, diabetes, and hypertension, respectively. The performance of the LLM-based pipeline for detecting conditions varied: AMI had 88% sensitivity, 63% specificity, and 77% positive predictive value (PPV); diabetes had 91% sensitivity, 86% specificity, and 71% PPV; and hypertension had 94% sensitivity, 32% specificity, and 72% PPV. Compared with ICD codes, the LLM-based method demonstrated improved sensitivity and negative predictive value across all conditions. The monthly percentage trends from the detected cases by LLM and reference standard showed consistent patterns.

**Conclusion**

The proposed LLM-based pipeline demonstrated reasonable accuracy and high efficiency in disease detection for multiple conditions. Human expert knowledge can be integrated into the pipeline to guide EHR note analysis without manually curated labels. The method could enable comprehensive real-time disease surveillance using EHRs.

## Introduction

Disease detection is a foundation for numerous critical preventive medicine and public health operations. It allows real-time monitoring of disease prevalence and incidence[1], [2], provides cohort identification for observational and interventional research[3], and enables novel patient stratification for precision medicine[4]. Detecting multiple diseases simultaneously from healthcare data is challenging since many conditions present with similar symptoms or have heterogeneous and ever-shifting presentations [5]. Electronic health records (EHRs) capture clinical decisions and orders by physicians, allied health providers, and system information throughout patient admission encounters. With the availability of EHR data, many artificial intelligence (AI)-based methods are increasingly being adopted to detect various conditions, including diabetes[6], [7], hypertension[8], cerebrovascular diseases[9], and adverse events[10], [11].

Recent AI-based methods have shown promising abilities in detecting medical conditions with precision [12]. However, the path to integrating these technologies into healthcare systems highlights several areas requiring improvement. For instance, due to its labour-intensive nature, manually labelling data for training models restricts the diversity of patient data and narrows model applications to limited cohorts [13], [14]. Besides, while effective for targeted diagnostics, the widespread focus on single-condition models risks oversimplifying the complex nature of patient health, where comorbidities are common. Considering the vast number of recognized diseases (over 9,000[15]) and potentially many yet undiscovered, it is neither practical nor efficient to train models using labels for every condition [4]. There is a compelling need for innovative approaches that can analyze complex, multi-condition, diverse patient data without relying on traditional labelling. This evolution requires not only technological innovation but also a collaborative effort among AI researchers, clinicians, and data scientists to create more adaptable, comprehensive, and effective healthcare solutions. Generative large language models (LLMs) provide a unique opportunity for developing such solutions[16].

Generative LLMs represent a transformative advancement in AI, trained to understand and align with human instructions in text and language analysis[17]. LLMs offer a few key strengths: 1) clinical expert knowledge can be integrated into prompts for explainable outcomes; 2) they can be applied to multiple clinical tasks[17]; 3) they exhibit a strong reasoning capacity in understanding text and human instructions; 4) they do not require reference labels, as in supervised learning models. If clinical expert knowledge is integrated correctly, LLMs hold the potential to support many medical applications, such as medical diagnosis[18], clinical criteria extraction[19], cancer treatment response[20], clinical image interpretation[21], treatment decision support[22], and medical education[23], [24].

Many existing studies have successfully applied generative LLMs to disease identification from EHR notes, including cancer stages [25], Severe Acute Respiratory Syndrome Coronavirus 2 [26], Alzheimer's disease [27], Parkinson's disease [28], and rare diseases [29], [30]. Most of the work focuses on using specific document types (e.g., pathology reports, patient's reported notes), cloud-based models (e.g., ChatGPT), and detecting specific conditions. A holistic view of how to detect multiple conditions from real-world EHR documents using localized generative LLMs is still lacking.

We aimed to design a practical pipeline to analyze and process a full range of EHR clinical notes for multiple condition detection. The pipeline combined LLMs with human clinical expertise to extract clinical diagnosis and disease management information. It was applied to the detection of acute myocardial infarction (AMI), diabetes, and hypertension, conditions known for their widespread prevalence worldwide and substantial impact on healthcare [31]. This pilot project provided a holistic view of evaluating localized LLM for multi-condition detection without human labelling but with expert knowledge, offering a scalable solution across healthcare systems.

## Materials and methods

### Pipeline architecture

We designed a pipeline integrating clinical expert knowledge and LLMs, as illustrated in Figure 1. The pipeline was used to determine disease status by input text and prompts based on human instructions without needing gold standard disease labels. The pipeline consisted of four components: 1) an LLM-based preprocessing of clinical notes to filter irrelevant information, 2) a prompt design for the LLM to infer disease presence status, 3) a text inference by the LLM with optimal hyperparameters, and 4) a post-processing of model responses with clinical rules. Disease status, i.e., present or absent, was determined for each patient through the pipeline.

*Figure 1 The architecture of the proposed pipeline for disease identification. The patient's raw electronic health records (EHRs) clinical notes went through the pipeline: data preprocessing, prompt design, text inference by the large language model (LLM), and post-processing for LLM's inference and information extraction results with clinical rules; the disease status was then determined. Human expert knowledge was applied to each component.*

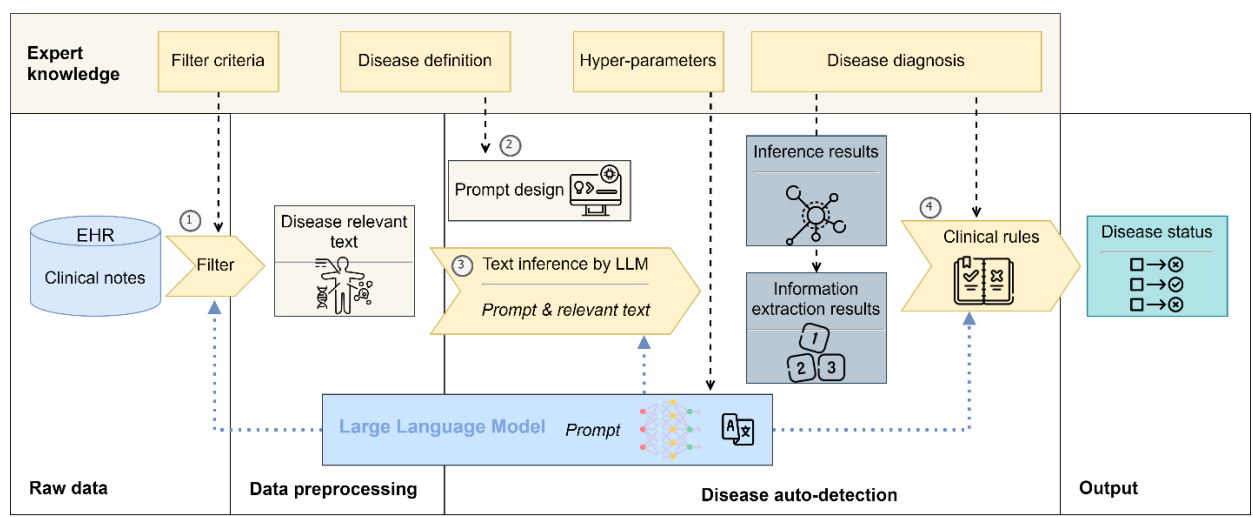

### Data source and reference labels

The study cohort consisted of patients admitted to hospitals in 2015 from the Clinical Registry, AdminisTrative Data and Electronic Medical Records (CREATE) database system, which contained the Alberta Provincial Project for Outcome Assessment in Coronary Heart Disease (APPROACH) clinical registry database linked to 3 other databases (SCM EHR, discharge abstract database, and national ambulatory care reporting system)[32]. The databases comprised patient demographics data, EHR clinical notes, and verified clinical diagnoses of the conditions of interest, including AMI, diabetes, and hypertension.

The reference labels of the three conditions were derived from the APPROACH clinical registry. APPROACH captures detailed procedures, demographic, and clinical information on all patients undergoing cardiac catheterization in Alberta. The data were collected by laboratory personnel through direct inquiry into patients and the procedure physician and review of medical documentation on clinical risk factors, including hypertension, diabetes, and a history of acute myocardial infarction [33]. The diagnosis data was then combined with diagnosis results from administrative health data using the International Classification of Diseases 10[th] version Canadian modification (ICD-10-CA) [34]. This study was approved by the Conjoint Health Research Ethics Board at the University of Calgary (REB19-0088 and REB-23-0535).

**Prompt-based data preprocessing**

In the extracted EHR data, each patient could have over 60 types of documents, such as nursing notes, discharge summaries, and clinical records, with some patients having multiple occurrences of many types. However, not all records contained information that was useful for condition identification. Including all clinical note types would add noise and consume extra computing resources, hampering model performance[35].

We designed a prompt-based data preprocessing method to extract the minimum sufficient text for LLM analysis. It consisted of four steps: document type sampling, document type inference, document type filtering, and document content selection, as shown in Figure 2. The first three steps involved analyzing the document types and eliminating those unrelated to the conditions of interest, and the final step filtered the notes within the selected types by extracting the most relevant sentences for each record.

*Figure 2 Prompt-based preprocessing of electronic health records (EHRs) notes. This process consists of four steps: document type sampling, document type inference, document type filtering, and document content selection. LLM stands for large language model. AMI stands for acute myocardial infarction.*

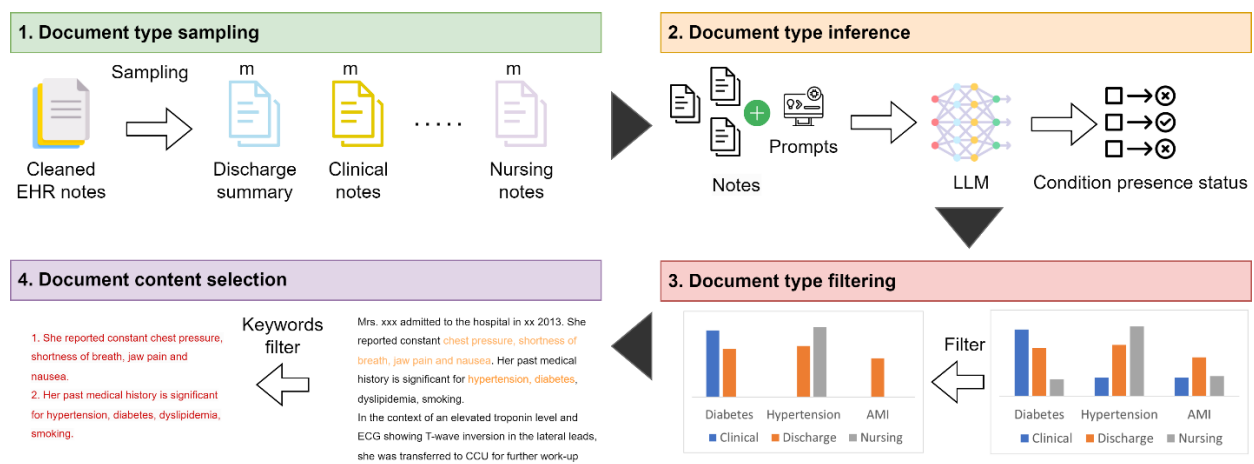

Sampling a manageable subset of notes helped focus on a representative sample while maintaining computational feasibility. For document type sampling, we randomly checked out $m$ records from each document type across all EHR notes. For document types with fewer than $m$ records, we included all available records. In documentation inference, each sampled record was sent to LLM for condition inference, where the prompt design was introduced in the following section. The results of LLM-based inference were not used for final disease detection but to calculate the Information Relevance (IR) score per document type in identifying cases. The IR is defined as the number of records having positive cases over $m$, as shown in Figure 2(3). During documentation type filtering, we set a threshold (larger than the $25^{th}$ percentile) to find important document types and keep only these types in the study notes. Patients lacking these document types were considered condition-free.

The subsequent analyses, such as keyword-based matching and disease detection, were conducted on the dataset after removing unnecessary document types. Keywords-based matching is to select useful chunks of text within the remaining documents per patient, as shown in Figure 2(4). These keywords were selected by a clinical expert (KR) and consisted of medical terms, such as "symptoms", "treatment", "medication", and specific terms for treatment management related to the targeted condition (see Table A. 1). The keywords were automatically searched in each document; the sentences which contained them were extracted and merged into a new document. The merged documents served as the final input for the subsequent disease detection by LLMs.

**Prompt design for disease identification**

Prompt design is the systematic practice of creating well-structured and effective instructions that guide LLMs to generate accurate, coherent, and contextually relevant outputs[36]. For disease identification, expert knowledge can be combined with a prompt to guide LLM in analyzing text and reducing model uncertainty. In real-world healthcare systems, healthcare professionals are trained to review clinical notes by identifying specific diagnoses, disease management information, and key clinical indicators to code conditions in administrative data. We designed prompts to retrieve this information for each condition to infer disease status.

Two prompting techniques were available in practice: 1) Information extraction, creating prompts to extract different types of information (e.g., symptoms and laboratory tests) and comparing the extracted results with clinical guidance for condition presence; or 2) Inferential reasoning, creating a prompt that infers the condition presence status by finding diagnosis evidence internally. In this study, we mainly used the second strategy for condition identification.

We referred to specific diagnoses, treatment management, and relevant clinical guidelines per condition to design prompts and clinical rules. In Canada, the clinical diagnosis of diabetes includes hemoglobin A1c levels of 6.5% or higher or glucose levels of 11.1 mmol/L [37]. AMI diagnosis involves characteristic electrocardiogram changes and elevated cardiac biomarkers such as troponin levels above the 99th percentile of the upper reference limit[38]. Hypertension is diagnosed as persistently elevated blood pressure readings, typically ≥140/90 mmHg, confirmed by multiple measurements[39].

The specific prompt templates for the three conditions are shown in Table 1. To supplement detection, we created an information extraction type of prompt to extract a key laboratory test per condition. A specific key laboratory test is often utilized in care systems in actual healthcare settings[40]. For example, we extracted the random blood sugar level for diabetes and compared it with a clinical standard of random plasma glucose ≥ 11.1 mmol/L[37]. Blood pressure was extracted for hypertension. The prompt for AMI was slightly different because many abbreviations were used in clinical notes. Though troponin levels were measured for AMI, electrocardiography was the standard test in Canada.

*Table 1 Prompt template for the identification of acute myocardial infarction (AMI), diabetes, and hypertension.*

| Conditions | Inferential reasoning type | Information extraction type |
|---|---|---|
| **AMI** | Analyze the clinical text: '{text}', and answer yes or no if you identify acute myocardial infarction. Be careful with some abbreviations for acute myocardial infarction, including ami, mi, stemi, and non-stemi. | Find all the key-value pairs of troponin level from the given text: {text}. |
| **Diabetes** | Analyze the clinical text: '{text}', answer yes or no if you identify diabetes. Look for relevant information, including elevated blood glucose levels, mentions of diabetes diagnosis, or references to anti-diabetic medications. | Find all the key-value pairs of blood sugar/glucose levels from the given text: {text}. |
| **Hypertension** | Analyze the clinical text: '{text}', answer yes or no if you identify hypertension (high blood pressure). Look for relevant information, including high blood pressure readings or symptoms, mentions of hypertension diagnosis, or references to antihypertensive medications. | Find all the key-value pairs of blood pressure from the given text: {text}. |

Notes: '{text}' refers to the placeholder of the real clinical notes.

**Generative large language models for text inference**

Clinical notes contain a high degree of complexity and variations, including specialized medical terminologies, abbreviations, and jargon. They are also a hybrid of semi-structured (i.e., key-value pairs) and unstructured formats recorded by different care providers and institutions. With these difficulties, the generalizability of traditional natural language processing (NLP) methods is limited. It is common that a pre-trained model may not perform well on another dataset without fine-tuning or retraining. With one model, the generative LLM could achieve robust generalizability across different tasks[19]. LLMs could parse, interpret, and summarize complex medical information in clinical notes data and transform unstructured text into structured knowledge that supports clinical decision-making[16], [19], [22], [24].

We implemented a state-of-the-art generative LLM, Mistral-7B-OpenOrca[41], [42], which obtained the best inference performance among models with 7 billion parameters. The model had already been pre-trained on a vast amount of high-quality text data[42], enabling the generation of accurate answers with a precise prompt. It was run in a secure and isolated health authority (i.e., Alberta Health Services) approved environment, using an NVIDIA Tesla V100 graphics processing unit (GPU). Its optimal hyperparameters were refined by grid search and given in Table A. 2.

The model was then used to analyze the medical terminologies and context and extract medical history, symptoms, laboratory results, diagnoses, and treatments from the notes. The inputs were each patient's preprocessed documents. The text within each document was inserted into the prompt template (see Table 1), serving as LLM inputs.

**Rule-based classification**

The rule-based classification is the postprocess of LLM responses. For each disease identification task, the LLM outputted two types of responses independently according to the prompts: a) inference response was the directly inferred presence status containing "Yes", "No", or "No mention", and b) information extraction response consisted of the key-value pairs of the laboratory test results, as shown in Table A. 3.

Each patient's clinical document was sent to LLM for disease inference. For type a) response, the rule was to check if "Yes", "No", or "No mention" was contained in the response and assigned the corresponding results to the document. For type b) response, if key-value pairs of laboratory tests were extracted for this document, we compared the results with clinical guidelines. For diabetes detection, the key was blood sugar level, and we checked if the value was larger than 11.1 mmol/L [37] to assign the "Yes" or "No" label to the document. Similarly, for hypertension, we compared the blood pressure (BP) with a standard if the mean 24-hour systolic BP was ≥ 140 mm Hg or diastolic BP was ≥ 90 mm Hg [39]. For AMI, we checked for a troponin level > 14 ng/L [38], [43]. If no test results were found in the document, a "No mention" was assigned. Eventually, we combined the labels for each document and merged these results as "Yes" or "No" per patient.

**Statistical analysis**

We calculated the prevalences of AMI, diabetes, and hypertension in the study cohort. Sex and comorbidities were calculated as ratios (%), and age and length of hospital stay were calculated as the median (interquartile range [IQR]). For document preprocessing, we calculated the percentage of records that contained relevant text in detecting the three conditions across all document types, respectively. We compared the inference results of the three conditions from the proposed pipeline against the reference standards[33] and compared the performance with the widely adopted ICD-10 codes-based detection methods[44]. The sensitivity, specificity, positive predictive value (PPV), and negative predictive value (NPV) per condition were reported. Their formula is given as follows,

$$Sensitivity = \frac{TP}{TP+FN}, \tag{1}$$

$$Specificity = \frac{TN}{FP+TN}, \quad (2)$$

$$PPV = \frac{TP}{TP+FP}, \quad (3)$$

$$NPV = \frac{TN}{TN+FN}, \quad (4)$$

where TP refers to true positive, FN refers to false negative, FP is false positive, and TN is true negative.

To visually assess the surveillance performance, the monthly percentage trends of the three conditions were graphed using the pipeline detection results and reference standards.

## Results

### Characteristics of the study participants

We obtained 3,088 patients from the CREATE database by selecting all the inpatients admitted in 2015. Their characteristics are shown in Table 2. Hypertension had the highest prevalence (65.9%), followed by AMI (55.4%) and diabetes (27.7%). Most patients were male (70%), ranging from 68.8% to 70.7% for three conditions. Patients with diabetes showed the highest prevalence of congestive heart failure (20.5%) and cerebrovascular diseases (7.0%). In contrast, the prevalence of chronic obstructive pulmonary disease was distributed across the three groups. There were also patients with multiple conditions of interest: a total of 714 (23.1%) patients having both diabetes and hypertension, 415 (13.4%) patients having diabetes and AMI, 1047 (33.9%) patients having hypertension and AMI, and 339 (10.9%) patients having all three conditions.

*Table 2 Characteristics of patient cohorts.*

|  | **Total** | **AMI** | **Diabetes** | **Hypertension** |
| --- | --- | --- | --- | --- |
| No. of patients (%) | 3088 (100) | 1710 (55.4) | 854 (27.7) | 2035 (65.9) |
| Male sex (%) | 2165 (70.1) | 1209 (70.7) | 592 (69.3) | 1400 (68.8) |
| Median age (IQR), years | 64.0 (55.8 - 72.8) | 63.3 (54.8 - 72.3) | 65.6 (58.5 - 73.2) | 66.2 (58.5 - 74.6) |
| Length of stay (IQR), days | 7.1 (2.0 - 8.0) | 6.3 (3.0 – 7.0) | 9.0 (3.0 - 9.2) | 7.2 (2.0 - 8.0) |
| **Comorbidities** | | | | |
| Congestive heart failure (%) | 455 (14.7) | 163 (9.5) | 175 (20.5) | 328 (16.1) |
| Chronic obstructive pulmonary disease (%) | 269 (8.7) | 117 (6.8) | 81 (9.5) | 191 (9.4) |
| Cerebrovascular disease (%) | 148 (4.8) | 72 (4.2) | 60 (7.0) | 126 (6.2) |

*AMI: Acute Myocardial Infarction; IQR: interquartile range.*

### Consolidation of clinical text

We gathered 551,095 EHR notes with 64 document types from the study cohort and then conducted prompt-based preprocessing to consolidate EHR notes. We sampled *m*=200 records per document type to determine the information relevance per condition. The calculated percentages of records containing relevant texts detecting the three conditions are shown in Figure 3. The common document types for the three conditions were "DischargeSummary", "CardiacDischarge", and "TransferSummary". AMI was less frequently documented in EHR notes than in the other two conditions. The detailed mapping between the document abbreviations and document names is reported in Table A. 4. Some documents were unique to the specific condition, such as "PatientCare" and "EDHandover" for AMI, "OutpatientConsultR" and "GoalFlowsheet" for diabetes, and "GoalAssessment" and "AlcoholAssessment" for hypertension. There were 20 document types that did not document any of these conditions. Figure 3 shows that many EHR notes included unnecessary text per condition.

*Figure 3 Information relevance of electronic health records (EHRs) document types for condition identification. Information relevance was the percentage of inferred positive diagnoses by the large language model (LLM) per document type. It is calculated for three conditions, such as acute myocardial infarction (AMI), diabetes, and hypertension.*

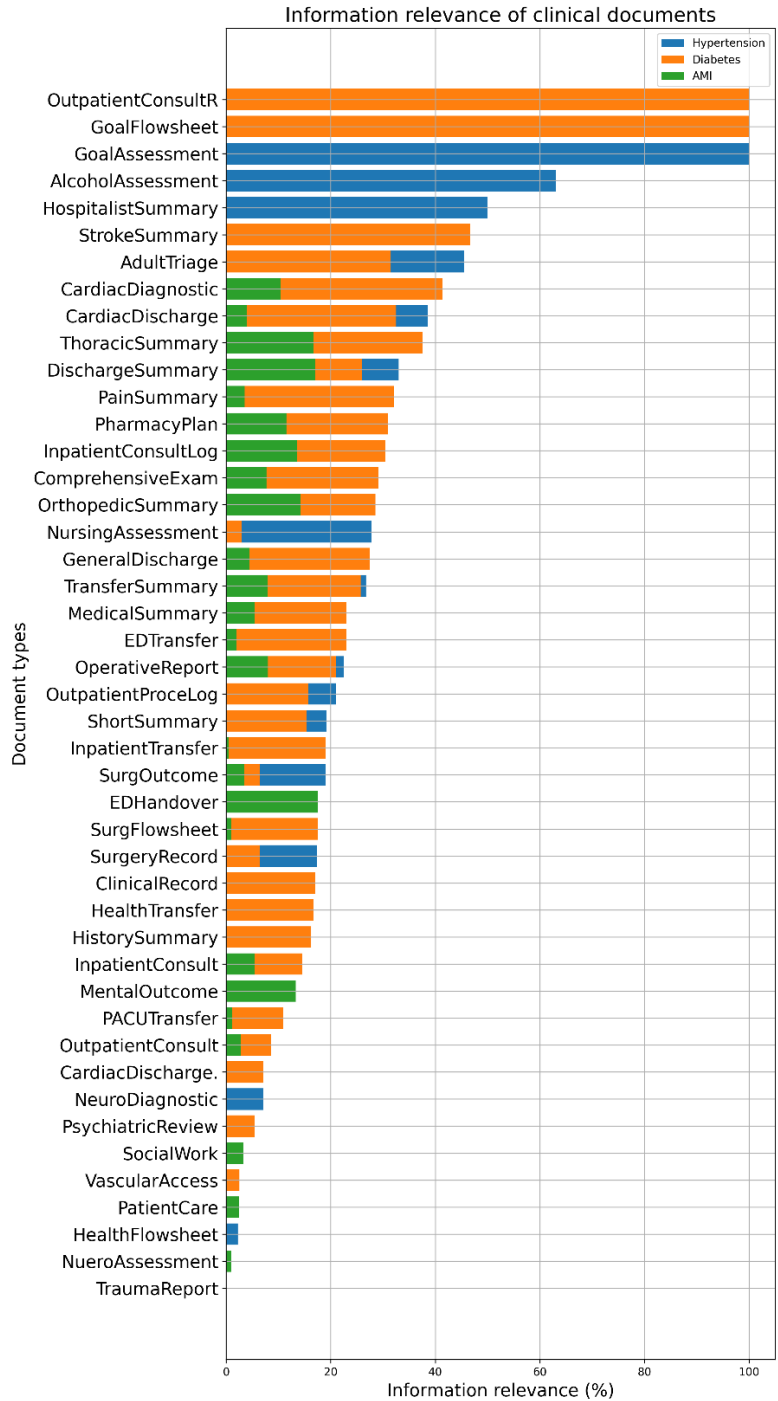

Running a single prompt over all the clinical documents without filtering took around 20 hours in our local settings. Considering we had two prompts for each of the three conditions, it was computationally costly. We had to preprocess the EHR notes to reduce the amount of input text sent to the pipeline. The 25th percentile (Q1) threshold used in document-type filtering in Figure 2(3) controls the percentage of filtered information. We had a comparative analysis of different levels of this threshold, such as 0, Q1, and Q2, as shown in Table A. 5. With an increase in the threshold level, the filtered document types and words increased, and the positive cases dropped for all three conditions. We chose Q1 to maintain a balanced filtering rate and the portion of positive cases retained.

After applying the 25th percentile threshold, there were 27 remaining document types for diabetes, 31 for hypertension, and 19 for AMI, as shown in Table 3. We then applied keyword searching to match sentences and merged them into a new document. Eventually, 70% of words in the entire cohort's notes were excluded for diabetes, 62% were excluded for hypertension, and 93% were removed for AMI (see words percentage in Table 3). The word percentage remaining is the ratio between the number of words in the whole dataset after and before text preprocessing. To examine whether the process removed sensitive information necessary for detecting conditions, we calculated the positive cases in the preprocessed text and found that at least 93% of patients with conditions were retained, indicating that essential information was preserved. We calculated the portion of positive cases retained by comparing patients with positive cases after and before text preprocessing.

*Table 3 Information consolidation after preprocessing for three conditions.*

| Condition | # document types (total N=64) | Words percentage remaining (%) | Portion of positive cases retained (%) |
| --- | --- | --- | --- |
| AMI | 19 | 6.9 | 93.3 |
| Diabetes | 27 | 29.7 | 98.5 |
| Hypertension | 31 | 37.9 | 97.9 |

AMI: Acute Myocardial Infarction.

## Explainability of LLM inference

Our strategies for disease identification were to look for specific diagnoses, disease management information, and clinical indicators from EHR notes. To examine if the LLM could find this information when making decisions, we displayed three response examples of LLM in Error: Reference source not found "Response to inference prompt". We classified the LLM responses into two main categories: direct and inferred diagnosis based on clinical context. In the direct diagnosis category, the LLM identified conditions explicitly mentioned in the clinical text, such as its response for AMI: "Yes, the text identifies acute myocardial infarction (AMI) as the patient has been diagnosed with AMI". This demonstrated the LLM's ability to recognize diagnoses that were clearly stated in the clinical notes.

In the inferred diagnosis based on the clinical context category, the LLM inferred the presence of a condition by interpreting relevant clinical information, such as medication use or test results, even when the diagnosis was not directly mentioned. For instance, in the case of diabetes, the LLM states: "Yes, you can identify diabetes in this clinical text. The text mentions critically high blood sugars and the administration of insulin (humulin r)". Conversely, the LLM also demonstrated caution by providing a negative response when no clear evidence was present, as seen in the hypertension example: "No, there is no clear mention of hypertension or high blood pressure in the given clinical text". These two categories reflect the LLM's diagnostic reasoning capabilities.

To further evaluate the LLM's context understanding and ability to provide introspection into its decision-making, we introduced an instruction in the prompts to "highlight all the original text that supports your judgement". An example of AMI detection is shown in Figure 4, where the LLM produced a positive inference and highlighted key supporting sentences, including "acute onset of chest pain", "diagnosed with an acute ischemic stroke", "troponins escalated", and "chest pain resolved". These sentences cover the symptoms, diagnosis, treatment, and lab tests for AMI, aligning with our identification strategies. This ability to trace its conclusions back to specific text segments improves the transparency and explainability of LLM-driven disease detection.

*Figure 4 Large language model (LLM) inference for acute myocardial infarction (AMI) detection based on a clinical note. The marked sentences were highlighted by the LLM in support of the presence of AMI.*

> **An example of LLM inference for AMI**
>
> Admission Date: [DATE], Discharge Date: [DATE] from [HOSPITAL].
> The patient, a [AGE]-year-old male, was brought to hospital by ambulance following ==acute onset of chest pain== radiating to his left arm, sweating, and nausea. Diagnosed ==with an acute myocardial infarction==. An ==ECG revealed anterior wall ST-elevation, and troponins escalated from 12 ng/L to 240 ng/L within 4 hours==.
> Immediate management with IV morphine, aspirin, and a beta-blocker alleviated the pain, and he was promptly moved to the cath lab. A coronary angiography showed a 90% blockage in the left anterior descending artery, which was successfully treated with primary percutaneous coronary intervention (PCI) and a drug-eluting stent placement. ==Post-PCI, his chest pain resolved==, and blood flow re-established...
> At discharge, the patient had stable condition with no chest pain and normal sinus cardiac rhythm. Vital signs were within normal range. Treatment plan included managing hypertension, hyperlipidemia, type 2 diabetes, and overweight ...

## Model Selection

To ensure the robustness of the selected LLM, Mistral-7B-OpenOrca, we evaluated several state-of-the-art LLMs for disease identification. We compared Mistral-7B-OpenOrca with three other models—OpenBioLLM-8B, Phi3-mini, and BioMedLM—on 10 predefined benchmarking questions (see Table A.6). The comparison results are shown in Table 4. OpenBioLLM-8B, a Llama-8B model fine-tuned on a medical and life science dataset, demonstrates performance comparable to Mistral-7B-OpenOrca but requires more processing time. Phi-3.5-mini-instruct achieves the best performance among the evaluated models; however, it is 17 times slower than Mistral-7B-OpenOrca in answering all benchmark questions. In contrast, BioMedLM performs the least effectively, providing only one correct answer. Mistral-7B-OpenOrca demonstrated competitive accuracy while using significantly less computational resources. Eventually, we selected Mistral-7B-OpenOrca as the final model for our pipeline due to its balanced performance and our resource-constrained environments where data privacy and computational efficiency are critical.

*Table 4 Benchmarking results of different LLMs on 10 predefined disease detection questions.*

|                       | Correct answers (%) | Time used (seconds) |
|-----------------------|---------------------|---------------------|
| Mistral-7BOpenOrca    | 80%                 | 31s                 |
| OpenBioLLM            | 70%                 | 63s                 |
| Phi-3.5-mini-instruct | 90%                 | 552s                |
| BioMedLM              | 10%                 | 27s                 |

## Model Performance

The detection performance of the three conditions is shown in Table 5. We compared the ICD-10 codes, the pipeline using the first prompt, the pipeline using the second prompt, and the pipeline with merged responses. The pipeline with merged responses performed similarly to the first prompt alone, with a slight improvement in detecting hypertension. The sensitivity of the second prompt was lower, likely because lab results are typically not documented in clinical notes. The cases detected by the second prompt largely overlapped with those identified by the first prompt.

The pipeline generally demonstrated high sensitivity (91%) and specificity (86%) for diabetes, although its positive predictive value (PPV) was lower (71%). For hypertension, the sensitivity was exceptional (94%), albeit with lower specificity (32%). Despite this, its PPV remains relatively high (72%). For AMI, the method demonstrated good sensitivity (88%) and specificity (63%), with a higher PPV of 77%.

The ICD-10 codes were part of the reference standards since they were included in the APPROACH clinical registry database as per its data collection protocols. For AMI, ICD-10 method achieved a sensitivity of 84% and a specificity of 87%, indicating a reliable balance between detecting true positive cases and true negative cases. For Diabetes, ICD-10's sensitivity was 91% and specificity was 97%. However, for hypertension, while the specificity remained high at 92%, the sensitivity was lower at 80%.

Combining LLM and ICD-10 generally improved sensitivity but led to a decrease in specificity. For AMI, the combined approach achieved the highest sensitivity at 95%, though specificity dropped to 60%. For diabetes, the combined sensitivity reached 94%, an improvement over either method alone, with a specificity of 85%. However, in the case of hypertension, while sensitivity was 96%, specificity was lower at 31%.

*Table 5 Comparison of detection performance for three conditions using physician labels as standard references.*

|  | **Condition** | **Sensitivity (%) (95% CI)** | **Specificity (%) (95% CI)** | **PPV (%) (95% CI)** | **NPV (%) (95% CI)** |
|---|---|---|---|---|---|
| **ICD-10 (Part of reference standard)** | AMI | 84.0 (83.8-84.1) | 87.0 (86.8-87.1) | 89.4 (89.3-89.6) | 80.6 (80.5-81.0) |
|  | Diabetes | 91.2 (91.0-91.3) | 97.9 (97.8-98.0) | 94.3 (94.2-94.5) | 96.8 (96.7-96.9) |
|  | Hypertension | 80.0 (79.1-81.2) | 92.3 (92.0-92.5) | 95.0 (94.9-95.1) | 71.4 (71.1-72.0) |
| **Pipeline (first prompt)** | AMI | 88.3 (88.2-88.4) | 63.4 (63.3-63.5) | 76.5 (76.4-76.7) | 80.0 (79.9-80.1) |
|  | Diabetes | 91.0 (90.0-91.1) | 85.5 (85.3-85.6) | 70.5 (70.4-70.6) | 96.1 (96.0-96.2) |
|  | Hypertension | 94.0 (93.9-94.1) | 32.1 (32.0-32.2) | 71.8 (71.7-71.9) | 74.3 (74.1-74.4) |
| **Pipeline (second prompt)** | AMI | 0.1 (0.1-0.1) | 99.9 (99.9-99.9) | 50.0 (47.8-52.2) | 41.5 (41.5-41.6) |
|  | Diabetes | 1.5 (1.4-1.5) | 100.0 (99.9-100.0) | 91.7 (91.2-92.2) | 73.5 (73.5-73.6) |
|  | Hypertension | 25.9 (25.9-25.9) | 83.6 (83.5-83.7) | 74.6 (74.5-74.7) | 37.8 (37.8-37.9) |
| **Pipeline (merged responses)** | AMI | 88.3 (88.2-88.4) | 63.4 (63.3-63.5) | 76.5 (76.4-76.7) | 80.0 (79.9-80.1) |
|  | Diabetes | 91.0 (90.0-91.1) | 85.5 (85.3-85.6) | 70.5 (70.4-70.6) | 96.1 (96.0-96.2) |
|  | Hypertension | 94.3 (94.2-94.3) | 32.6 (32.5-32.7) | 72.2 (72.1-72.2) | 75.5 (75.3-75.6) |
| **Pipeline + ICD-10** | AMI | 94.9 (94.8-95.0) | 60.4 (60.2-60.5) | 76.6 (76.4-76.6) | 89.7 (89.5-89.7) |
|  | Diabetes | 93.6 (93.5-93.6) | 85.2 (85.1-85.3) | 70.1 (70.0-70.2) | 97.3 (97.2-97.3) |
|  | Hypertension | 96.2 (96.2-96.3) | 32.2 (32.1-32.3) | 72.4 (72.3-72.4) | 81.9 (81.7-82.0) |

AMI: Acute Myocardial Infarction; CI: confidence interval.

To test the performance of the proposed method over time, we evaluated the monthly percentages of the three conditions identified by the proposed method and physician diagnosis, as shown in Figure 5. For AMI, diabetes, and hypertension, the predicted numbers closely aligned with the changes in actual cases, indicating a consistent performance throughout the year with minor deviations that did not significantly affect the overall trend.

*Figure 5 The comparison of monthly percent of cohort with disease identified by the proposed pipeline and physician diagnosis. "Predicted *" stands for the results by the pipeline, otherwise by physician diagnosis.*

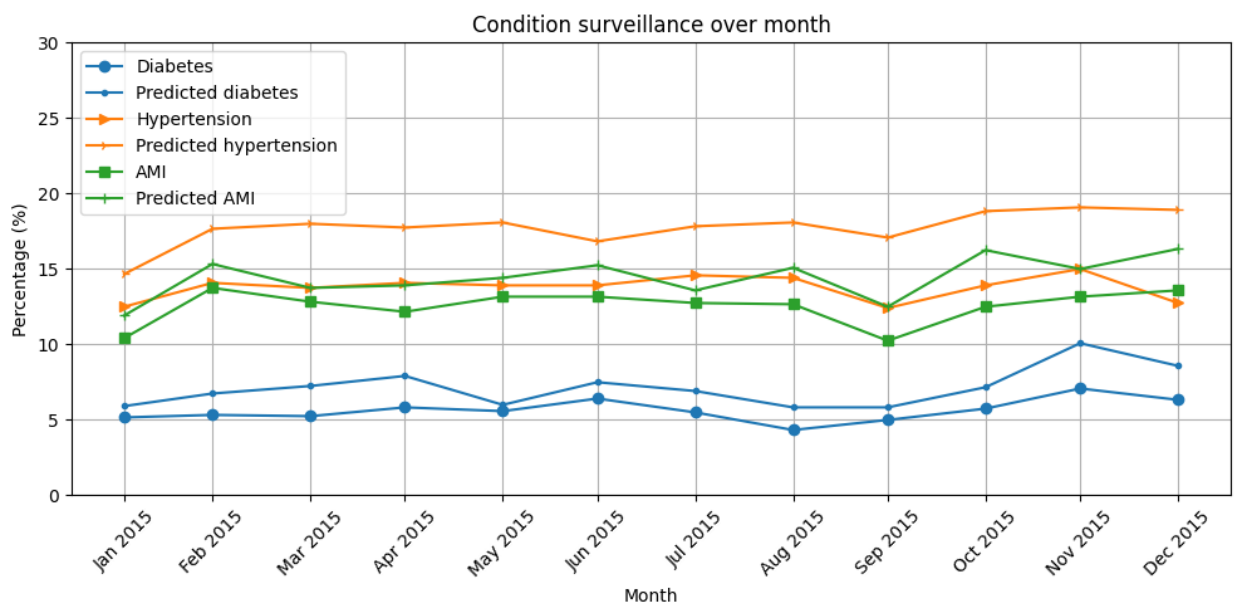

**Discussion**

This study proposed an innovative pipeline to detect multiple conditions by automatically analyzing large-scale EHR notes without the need for manual data labelling. The pipeline, which integrated an LLM with human instructions from expert knowledge, demonstrated a reasonable performance in identifying conditions such as AMI, diabetes, and hypertension. The approach yielded high sensitivity and NPV across the conditions compared to widely adopted ICD-10 methods. The monthly detected percentages closely mirrored the pattern of actual diagnoses, exhibiting as an efficient real-time surveillance indicator for healthcare outcomes.

Notably, the ICD-10 codes used in our study are part of the reference standards and are merged with APPROACH labels at the time of catheterization, so the high performance of ICD codes is expected here. Some studies reported the performance of ICD-10 in regular settings, where ICD codes are not involved in the reference standards. For instance, the sensitivity of ICD-10 was 47-78% for hypertension [8], [44], 72% for AMI [44], and 63-84% for diabetes [6], [44]. Higher sensitivities could be expected in this cohort as the patients were cardiac-specific, and these conditions would be germane to their visit. The combined LLM and ICD approach demonstrated improved sensitivity and NPV across all conditions but showed lower specificity and PPV. This approach might be suitable in scenarios where identifying true cases is more critical than minimizing false positives; for example, to manage chronic diseases, the combined method or LLM alone could ensure that most cases with conditions are screened and identified, with a follow-up test to rule out false positives.

Existing disease detection methods rely either on administrative health data (e.g., ICD-10) algorithms or supervised machine learning methods. In Canadian settings, the ICD-based method relies on the quality of codes provided by nationally trained coders who review patient charts[45]. However, many conditions are under-coded due to increased information overload for coders and the healthcare system[45], causing low sensitivity in detecting conditions. Therefore, supervised machine learning methods are being developed on EHR data to enhance the detection performance. Supervised methods require manually labelled health outcomes, limiting their application to a large-scale population, diverse disease identification, and real-time surveillance. With the wide adoption of EHR systems, unlabelled health data is exponentially increasing globally, while labelled data can only grow linearly due to limited resources. The proposed pipeline could scale to handle high volumes of EHR data and detect various conditions at a population level.

AI methods are not systematically utilized in medical practice. One of the most important reasons is their unexplainable decision processes. As demonstrated in the proposed method, clinical expertise was included as part of the pipeline in plain text format, simplifying maintenance and troubleshooting. Unlike traditional NLP methods, the pipeline was able to use pieces of evidence to support transparent decisions (Figure 4). In addition, the pipeline could be run entirely within a healthcare institutional firewall, aligning with patient privacy and data security needs. The clear and transparent process and adherence to data privacy standards will likely enhance utilization considerations within the health industries.

Generally, we observed three patterns from the experiment results. First, sensitivity was higher than PPV (mean 91% vs 73% across the three conditions). This indicates that besides accurately including true positive cases, the LLM model could include cases that were not diagnosed as positive. Second, the low specificity was primarily due to false positive cases, which we analyzed across the three conditions. The false positives fall into several categories:

1. Specific diagnoses were mentioned in clinical notes but not labelled in the reference standards (accounting for 25% of total cases). For example, discharge summaries often mentioned specific diagnoses such as NSTEMI/STEMI for AMI, hypertension, and diabetes. The pipeline identified these as positive cases, which was reasonable but not reflected in the reference labels. The reference labels were collected by laboratory personnel through direct inquiry into patients and the catheterization procedure physician, and review of medical documentation, and then combined with diagnosis results from administrative health data using ICD-10 codes. However, the comprehensiveness of data collection varied across personnel.
2. Highly suspected conditions (40-50%): For AMI, clinical notes described chest pain and elevated or positive troponin levels. For hypertension, notes may document high blood pressure readings (e.g., "systolic 192, diastolic 102") and mention antihypertensive medications like Ramipril. For diabetes, notes may include hemoglobin A1c levels and medications like metformin. These cases may represent borderline cases with similar symptoms requiring the same medications. It is possible that the models had difficulty reliably distinguishing them.
3. History of the conditions (15%): Clinical notes may reference a patient's personal or family history of the conditions in question, which led to false positives when active diagnoses were not present.
4. Other cases (10-20%): For instance, the LLM sometimes classified acute coronary syndrome (ACS) as AMI, leading to misidentifications. In hypertension, the LLM occasionally misclassified normal blood pressure readings (e.g., 112/78) as positive hypertension, particularly when the notes contained a large number of blood pressure readings. For diabetes, medications like Synthroid (not used for diabetes treatment) were occasionally misinterpreted as indicators of the condition.

The reasons behind these include some positive cases that were not captured in the reference labels from the APPROACH clinical registry. The clinical registry's data collection process was more restrictive and subject to physician confirmation. For example, diagnosing hypertension is more complex, requiring multiple blood pressure measurements, whereas diabetes diagnosis is more straightforward and is often based on hemoglobin A1c levels. Furthermore, the LLM struggled to distinguish subtle nuances between conditions (e.g., ACS vs. AMI) due to similarity of the medical conditions and exhibited inconsistencies in mathematical comparisons, such as comparing the blood pressure values with diagnosis standards. Nevertheless, identifying suspected cases remained valuable, as these patients may be at risk for developing the condition, requiring earlier preventive interventions.

Lastly, the detection performance varied across conditions. Diabetes detection showcased superior results (91% sensitivity, 86% specificity, 71% PPV, and 96% NPV) compared to other conditions. This discrepancy might stem from varying documentation standards. In Alberta, diabetes coding in administrative data and EHR systems is compulsory, likely leading to higher coding quality than for other conditions. Accordingly, the pipeline could serve as an indicator of documentation quality for various medical conditions with varying detection performance if reference standards were provided. The higher performance, characterized by both high sensitivity and specificity, could indicate better document quality.

The performance of the pipeline varied with preprocessing, prompt design, LLM model, and post-processing. The EHR data included all clinical notes per patient during hospitalization, with an average of 178 notes per patient in this study. Our designed preprocessing strategy using prompts significantly reduced the text volume (75% word reduction on average) for LLM analysis, thereby improving efficiency and accuracy by minimizing noise. The prompts for preprocessing and condition detection were the same and were designed based on clinical diagnosis guidelines per disease. The LLM we implemented was a small-scale model due to memory limitations (16 GB) with the available GPU, which limited its capacity to process complex prompts. For example, you may need more specific prompts which include symptoms, laboratory tests, and medications instead of a broad and general one (e.g., "Does this patient have a diagnosis of diabetes given all the clinical notes?"). Providing a few examples with some annotated data would improve the retrieval of desired content.

We experimented with both overly detailed prompts that included clinical guidelines (e.g., troponin levels >14 ng/L for AMI, and specific medications taken) and moderate instructions, as reported in our study. Interestingly, the moderate prompts yielded better performance in detecting these conditions, likely because overly complex prompts can overwhelm the model or introduce ambiguity in decision-making. Performance is influenced by both the prompts and the specific model used. We believe that the optimal prompt may vary depending on the specific LLM, its training process, and the datasets being used. A few strategies could always refine the prompt, including iterative refinement, where prompt wording and structure are continuously tested and tweaked to improve model performance, and human-in-the-loop evaluation, which involves domain experts reviewing outputs to provide qualitative feedback for prompt adjustments.

Since the inference capabilities of LLMs vary according to data and architectures, evaluating them before usage is often necessary. We applied post-processing in the pipeline to handle extracted laboratory test results by comparing them with clinical guidelines. The post-processing could be further improved by constructing a logistic regression model based on laboratory test trajectories to adapt to diverse tasks. With the customization and optimization of these components, the pipeline could be extended to handle more healthcare questions.

Although we integrate human knowledge in prompt design to guide LLM's behaviours, it has several inherent limitations. One major challenge is that prompts are external instructions and cannot rectify fundamental limitations of LLMs, such as their baseline performance in reasoning and clinical knowledge understanding. Moreover, prompt design often requires extensive trial and error to create prompts that elicit accurate, reliable outputs, especially when dealing with complex or ambiguous medical data. This process can be labour-intensive and may not consistently yield improvements across all disease identification tasks. Additionally, prompts are typically limited in their capacity to handle clinical nuances, as they may struggle to capture the full range of variability in clinical presentations, medical terminologies, and clinical abbreviations. This makes it challenging to design prompts that consistently account for the ambiguity often present in medical data, potentially leading to disease detection or classification errors.

The study has several limitations. First, further examination of false positive cases is needed to improve the model performance for future implementations within clinical care settings. Since these cases could be highly suspected patients, a specific LLM fine-tuned on relevant data is beneficial to distinguish them from firmly diagnosed cases. Second, our existing computing infrastructure limited us from utilizing larger-scale LLMs with superior inference capacities. We plan to upgrade this infrastructure, enabling the application and fine-tuning of advanced LLM models. However, recent studies [46], [47] show that fine-tuning LLMs on biomedical data without appropriate strategies does not always lead to improved outcomes. It can sometimes reduce performance on clinical tasks compared to general-purpose models. Lastly, our pipeline was validated on a cardiac disease cohort in Calgary and has not been evaluated in external databases. We aim to collaborate with other institutions from multiple geographic regions in the future to strengthen the validity of the method.

## Conclusion

Our proposed pipeline demonstrates the feasibility of integrating human expert knowledge with LLMs to infer multiple diseases from EHR clinical notes at scale. By leveraging generative LLMs as foundational models, the proposed approach can detect multi-conditions and expand to a broader range of conditions with customized data preprocessing and specific prompts. We are poised to expand beyond the limitations of detecting only one or a few cases, offering a scalable and adaptable solution for complex, multi-condition diagnostics from real-time health data.

## Acknowledgement

The authors thank Dr. Alexander Leung for his invaluable suggestions and comments on the clinical knowledge usage in method development, improving the work's integrity.

## Funding sources

This work was supported by Canadian Institutes of Health Research Operating Project Grants (201809FDN-409926-FDN-CBBA-114817 for HQ and 202209PJT-486541-HS1-CBBA-68649 for NL).

## Availability of data and materials

The patient data, which underpin the conclusions of this study, cannot be shared due to their potentially identifiable contents and institutional data privacy policies.

## Appendices

*Table A. 1 Keywords used to select sentences for input text.*

| Condition | Keywords |
| --- | --- |
| AMI | age\|weight\|wt\|myocardial infarction\|myocardial\|heart\|mi\|acute coronary\|coronary\|ischemic\|cardiac\|myocardium\|infarct\|ecg\|troponin\|artery\|pci\|stemi\|nstemi\|cardiogenic\|aneurysm\|medication |
| Diabetes | age\|weight\|wt\|non-alcoholic fatty liver\|dyslipidemia\|sugar\|dyslipidemia\|hypertension\|blood pressure\|glycemia\|glucose\|fasting\|fpg\|ogtt\|hba1c\|a1c\|mmtt\|hemoglobin\|insulin\|diabetes\|diabetic\|dm\|tolerance\|inhibitor\|peptide \|tzds\|glp-1\|inhibitors\|dpp-4\|metformin\|medication |
| Hypertension | age\|weight\|wt\|hypertension\|blood pressure\|systolic\|diastolic\|htn\|dash\|hypertensive\|medication |

AMI: Acute Myocardial Infarction.

*Table A. 2 Optimized hyperparameters for LLM after grid search.*

| Parameters | Description | Optimal value (range) |
| --- | --- | --- |
| Temperature | The temperature parameter controls the randomness in the probability distribution used to select the next word or token in the generated text. A higher value increases the randomness and diversity; lower values produce more deterministic and predictable text. | 0.5 (0-1) |
| Top_p | A cumulative probability threshold to select the next word from a subset of the vocabulary. Higher values allow for a diverse but might less relevant selection; and vice versa. | 0.9 (0-1) |
| Top_k | Top_k sampling restricts the word selection to the k most likely next words, reducing the chance of selecting highly improbable words. Higher values introduce more variety but might bring less relevant selection, and vice versa. | 50 (>1) |

*Table A. 3 Input and response examples of LLM for disease identification.*

| Condition | Response to inference prompt | Response to information extraction prompt | Input text |
| --- | --- | --- | --- |
| AMI | Response: Yes, the text identifies acute myocardial infarction (AMI) as the patient has been diagnosed with AMI). | Response: troponin level: 1.16 ng/mL. | A 68-year-old male presented to ER after experiencing severe chest pain during a morning jog. He was diagnosis with acute myocardial infarction (AMI), supported by troponin I levels elevated at 1.16 ng/mL. Past medical history included hypercholesterolemia and hypertension for which he was on atorvastatin and lisinopril. After acute management, he was immediately transferred to a specialized cardiac care facility for further management… |
| Diabetes | Response: Yes, you can identify diabetes in this clinical text. The text mentions critically high blood sugars and the administration of insulin (humulin r), which is an anti-diabetic medication. | Response: Key-value pairs: 1. glucose - mmol/l breakfast: 24.8 mmol/l 2. glucose - mmol/l breakfast: 20.1 mmol/l 3. glucose - mmol/l breakfast: 16.6 mmol/l 4. poct blood glucose - mmol/l lunch: 12 mmol/l 5. glucose - mmol/l lunch: 9.7 mmol/l 6. glucose - mmol/l lunch: 8.9 mmol/l | glucose - mmol/l breakfast : mmol/l comment : critically high blood sugars now. insulin (humulin r), 21 units given now. weight kg : 54.9 kg height cm : 158 cm. glucose - mmol/l breakfast : 24.8 mmol/l. glucose - mmol/l breakfast : 20.1 mmol/l. glucose - mmol/l breakfast : 16.6 mmol/l poct blood glucose - mmol/l lunch : 12 mmol/l. glucose - mmol/l lunch : 9.7 mmol/l. glucose - mmol/l lunch : 8.9 mmol/l. glucose - mmol/l other : 6.5 mmol/l. glucose - mmol/l |

| | | 7. glucose - mmol/l other: 6.5 mmol/l<br>8. glucose - mmol/l other: 5.8 mmol/l<br>9. glucose - mmol/l other: 6.6 mmol/l<br>10. glucose - mmol/l other: 8.7 mmol/l<br>11. glucose - mmol/l other: 7.4 mmol/l<br>12. glucose - mmol/l other: 6.1 mmol/l<br>13. glucose - mmol/l other: 6.8 mmol/l<br>14. glucose - mmol/l other: 93 mmol/l<br>15. glucose - mmol/l other: 83 mmol/l<br>16. glucose - mmol/l other: 17 mmol/l<br>17. glucose - mmol/l other: 16.7 mmol/l<br>18. weight kg: 54.9 kg<br>19. height cm: 158 cm<br>20. insulin (humulin r): 20 units given now<br>21. blood pressure systolic: 140<br>22. blood pressure diastolic: 66<br>23. blood pressure mean: (not provided) | other : 5.8 mmol/l. glucose - mmol/l other : 6.6 mmol/l. glucose - mmol/l other : 8.7 mmol/l. glucose - mmol/l other : 7.4 mmol/l. glucose - mmol/l other : 6.1 mmol/l. glucose - mmol/l other : 6.8 mmol/l. glucose - mmol/l other : 93 mmol/l. glucose - mmol/l other : 83 mmol/l. glucose - mmol/l other : 17 mmol/l. glucose - mmol/l other : 16.7 mmol/l. blood pressure blood pressure systolic : 140 comment : after exertion of transferring from wheelchair to bed… |
| Hypertension | Response: No, there is no clear mention of hypertension or high blood pressure in the given clinical text. | Response: There are no key-value pairs of blood pressure in the given text. | Admission Date: [DATE], Discharge Date: [DATE] from [HOSPITAL]<br>The patient, a [AGE]-year-old male, presented to ER with acute onset of chest pain radiating to his left arm, sweating, and nausea…<br>An ECG revealed anterior wall ST-elevation, and troponins escalated from 12 ng/L to 240 ng/L within 4 hours. Diagnosed with acute myocardial infarction.<br>Immediate management with IV morphine [dose], aspirin [dose], and [a beta-blocker [dose]…. The patient was promptly moved to the cath lab. A coronary angiography showed a 90% blockage in the left anterior descending artery, which was successfully treated with primary percutaneous coronary intervention (PCI) and a drug-eluting stent placement. Post-PCI, chest pain resolved, and blood flow re-established ... |

|  | At discharge, the patient had stable condition with no chest pain and normal sinus cardiac rhythm. Vital signs were within normal range. Treatment plan included managing hyperlipidemia, type 2 diabetes, and overweight ... |
|---|---|

AMI: Acute Myocardial Infarction.

*Table A. 4 Mapping between document names and abbreviations.*

| Document name | Shorten names | Content description |
|---|---|---|
| Acute Pain Summary | PainSummary | A summary of a patient's acute pain evaluation and management. |
| Trauma Admission Report | TraumaReport | Details of the assessment and care provided during a trauma patient's hospital admission. |
| Adult Emergency Triage Note | AdultTriage | Initial assessment and prioritization of an adult patient in the emergency department. |
| Blood Product Reaction Log | BloodLog | Record of any reactions to blood product transfusions. |
| Cardiac Diagnostic Report | CardiacDiagnostic | Diagnostic findings related to the patient's cardiac function and conditions. |
| Clinical Encounter Record | ClinicalRecord | Documentation of a clinical visit or interaction between patient and provider. |
| Day Surgery Record | SurgeryRecord | Information on the patient's day surgery, including procedure and outcomes. |
| Cardiac Discharge Summary | CardiacDischarge | Summary of a patient's condition and instructions upon discharge after cardiac care. |
| General Discharge Summary | GeneralDischarge | Summary of the patient's hospital stay and discharge instructions. |
| Hospitalist Discharge Summary | HospitalistSummary | Discharge summary prepared by the hospitalist overseeing patient care. |
| Medical Discharge Summary | MedicalSummary | Summary of a patient's medical condition and treatment upon discharge. |
| Orthopedic Surgery Discharge Summary | OrthopedicSummary | Discharge details for a patient who underwent orthopedic surgery. |
| Stroke Neurology Discharge Summary | StrokeSummary | Discharge report summarizing care for a patient treated for a stroke. |
| Short Surgery Discharge Summary | ShortSummary | Summary of discharge for short or minor surgical procedures. |
| Thoracic Surgery Discharge Summary | ThoracicSummary | Discharge summary following thoracic surgery. |
| General Discharge Summary (Duplicate) | DischargeSummary | Summary of the patient's hospital stay and discharge. |
| ED Handover Report | EDHandover | Information passed during handover from emergency to inpatient care. |
| Goal Achievement Assessment | GoalAssessment | Assessment of a patient's progress towards health-related goals. |
| Goal Achievement Flowsheet | GoalFlowsheet | Flowsheet tracking progress towards goals in patient care. |
| Comprehensive History & Physical Exam | ComprehensiveExam | Detailed history and physical exam conducted for thorough patient evaluation. |
| History & Physical Summary | HistorySummary | A brief summary of the patient's history and physical examination. |
| Inpatient Consultation Report | InpatientConsultLog | Report documenting the consultation for an inpatient by a specialist or consultant. |
| Inpatient Consultation | InpatientConsult | Notes from a consultant reviewing an inpatient's condition. |
| Operative/Procedure Detail Report | OperativeReport | Detailed report on the surgery or procedure performed on the patient. |
| Psychiatric Review (General) | PsychiatricReview | General review of the patient's psychiatric status and treatment. |
| Med-Surg Outcome Assessment | SurgOutcome | Assessment of outcomes after medical or surgical treatment. |
| Med-Surg Outcome Flowsheet | SurgFlowsheet | Flowsheet tracking medical or surgical treatment outcomes. |
| Mental Health Outcome Assessment | MentalOutcome | Evaluation of patient outcomes in mental health treatment. |
| Mental Health Outcome Flowsheet | HealthFlowsheet | Flowsheet monitoring outcomes in mental health care. |
| Neuro Diagnostic Report | NeuroDiagnostic | Report detailing neurological diagnostic tests and results. |
| ED to Inpatient Transfer Note | EDTransfer | Documentation of transfer from emergency department to inpatient care. |
| Inpatient Transfer Note | InpatientTransfer | Notes documenting the patient's transfer between inpatient units. |
| Mental Health Transfer Note | HealthTransfer | Report on the transfer of a mental health patient to another facility or care unit. |
| PACU to Inpatient Transfer Note | PACUTransfer | Transfer note from post-anesthesia care unit (PACU) to inpatient unit. |
| Outpatient Consultation Report | OutpatientConsultR | Detailed report of an outpatient consultation. |
| Outpatient Consult | OutpatientConsult | Summary of an outpatient consultation and care plan. |
| Outpatient Procedure Note | OutpatientProceLog | Notes from an outpatient procedure including outcomes and follow-up. |
| Vascular Access Record | VascularAccess | Documentation of vascular access procedures performed on the patient. |
| Neurological Patient Assessment | NueroAssessment | Assessment of the patient's neurological status. |
| Patient Care Summary | PatientCare | General summary of patient care provided during hospitalization. |
| Pharmacy Treatment Plan | PharmacyPlan | Plan outlining the patient's medication therapy managed by the pharmacy. |
| Social Work Evaluation | SocialWork | Assessment conducted by a social worker regarding the patient's social needs. |
| Pre-Op Nursing Assessment | NursingAssessment | Pre-operative assessment completed by nursing staff. |
| Transfer Summary Note | TransferSummary | Summary of patient's condition and treatment during transfer to another facility. |
| Alcohol Withdrawal Assessment | AlcoholAssessment | Evaluation of the patient's symptoms and status related to alcohol withdrawal. |

Table A. 5 The comparison of different levels of threshold in document type filtering.

| Condition | Information relevance threshold | #Document types | Words percentage remaining | Portion of positive cases retained |
|---|---|---|---|---|
| Diabetes | > 0 | 36 | 0.42 | 0.986 |
| Hypertension | > 0 | 42 | 0.571 | 0.979 |
| AMI | > 0 | 25 | 0.16 | 0.942 |
| Diabetes | > Q1 | 27 | 0.297 | 0.985 |
| Hypertension | > Q1 | 31 | 0.379 | 0.979 |
| AMI | > Q1 | 19 | 0.069 | 0.933 |
| Diabetes | > Q2 | 18 | 0.065 | 0.974 |
| Hypertension | > Q2 | 21 | 0.134 | 0.952 |
| AMI | > Q2 | 11 | 0.044 | 0.862 |

Q1: 25$^{th}$ percentile;

Q2: 50$^{th}$ percentile.

Table A. 6 Benchmark questions to test different localized LLMs

| | Question | Answer |
|---|---|---|
| Q1 | Imagine you are a physician, does the following text contain lab tests used to detect sepsis: fasting plasma glucose (FPG) test, oral glucose tolerance test (OGTT), hemoglobin A1c (HbA1c) test, and random plasma glucose (RPG) test? | No |
| Q2 | Imagine you are a physician, does the following text contain lab tests used to detect diabetes: fasting plasma glucose (FPG) test, oral glucose tolerance test | Yes |
| | (OGTT), hemoglobin A1c (HbA1c) test, and random plasma glucose (RPG) test? | |
| Q3 | What is the systolic blood pressure from the given text: Temperature Degrees C 36.2 degrees CPulse Pulse bpm : 72 bpm Blood Pressure Blood Pressure Systolic : 119 Blood Pressure Diastolic : 71 Blood Pressure Mean : 87 mmHg Blood Pressure Patient Position? | Systolic: 119 |
| Q4 | The clinical note states: 'The patient has a history of high blood sugar and is currently on insulin therapy.' Can you identify if the patient has diabetes? | Yes |
| Q5 | 'The patient was diagnosed with hypertension 5 years ago and has been on lisinopril since. No signs of improvement. Can you extract the diagnosis of hypertension and recognize when it occurred? | Hypertension 5 years ago |
| Q6 | 'The patient was admitted with acute chest pain, later confirmed to be a myocardial infarction. They also have a long-standing history of hypertension and are managing diabetes with metformin.' Can you identify the three conditions: myocardial infarction, hypertension, and diabetes? | Myocardial infarction, diabetes, and hypertension |
| Q7 | 'Patient reported severe chest pain radiating to the left arm, with nausea and shortness of breath. EKG confirmed ST elevation.' Can you identify if this patient is likely suffering from an acute myocardial infarction based on the symptoms and test results? | Yes |
| Q8 | 'The patient is obese, with a family history of diabetes and hypertension. Fasting glucose levels are elevated, and blood pressure remains uncontrolled despite medication.' Based on the risk factors and medical history, can you infer the likelihood of diabetes and hypertension in this patient? | Highly likely that the patient has diabetes; almost certainly has hypertension. |
| Q9 | 'The patient is currently on metformin, atorvastatin, and hydrochlorothiazide.' Can you identify which conditions these medications are most likely treating? | Type 2 diabetes and hypertension |
| Q10 | 'The patient was evaluated for chest pain, but there is no evidence of myocardial infarction. He has diabetes but no signs of hypertension.' Can you correctly identify the presence of diabetes while acknowledging that there is no myocardial infarction or hypertension? | Yes, the patient has diabetes but no MI or hypertension. |